\documentclass[11pt,a4paper]{article}
\usepackage[hyperref]{acl2021}
\usepackage{times}
\usepackage{latexsym}

\usepackage{graphicx}

\usepackage{microtype}

\aclfinalcopy 

\title{Bilingual Language Modeling, A transfer learning technique for Roman Urdu}

\author{Usama Khalid \\
  AIM Lab, NUCES (FAST)\\
  Islamabad, Pakistan \\
  \texttt{usama.khalid@nu.edu.pk} \\\And
Mirza Omer Beg \\
  AIM Lab, NUCES (FAST)\\
  Islamabad, Pakistan \\
  \texttt{omer.beg@nu.edu.pk} \\\AND
Muhammad Umair Arshad\\
  AIM Lab, NUCES (FAST)\\
  Islamabad, Pakistan \\
  \texttt{umair.arshad@nu.edu.pk}\\}

\date{}
\begin{document}
\maketitle
\begin{abstract}
Pretrained language models are now of widespread use in Natural Language Processing. Despite their success, applying them to Low Resource languages is still a huge challenge. Although Multilingual models hold great promise, applying them to specific low-resource languages e.g. Roman Urdu can be excessive. In this paper, we show how the code-switching property of languages may be used to perform cross-lingual transfer learning from a corresponding high resource language. We also show how this transfer learning technique termed \textit{Bilingual Language Modeling} can be used to produce better performing models for Roman Urdu. To enable training and experimentation, we also present a collection of novel corpora for Roman Urdu extracted from various sources and social networking sites, e.g. Twitter. We train Monolingual, Multilingual, and Bilingual models of Roman Urdu - the proposed bilingual model achieves 23\% accuracy compared to the 2\% and 11\% of the monolingual and multilingual models respectively in the Masked Language Modeling (MLM) task. 

\end{abstract}

\section{Introduction}

\begin{table*}
\centering
\begin{tabular}{llll}
\hline
\textbf{Corpus} & \textbf{Reference} & \textbf{Sentence Count} & \textbf{Word Count}\\
\hline
Urdu NER & \cite{khananamed} & 1,738 & 49,021 \\
COUNTER & \cite{sharjeel2017counter} & 3,587 & 105,124 \\
Urdu Fake News & \cite{amjad2020bend} & 5,722	 & 312,427 \\
Urdu IMDB Reviews & \cite{azam2020sentiment} & 608,693 & 14,474,912 \\
Roman Urdu sentences & \cite{sharf2018performing} & 20,040 & 267,779\\
Roman Urdu Twitter & Proposed & 3,040,153 & 54,622,490\\
\hline
\end{tabular}
\caption{ Statistics of the collected Urdu and Roman Urdu corpora. The Urdu corpora have all been cleaned and transliterated to Roman Urdu. In addition to this a novel corpus for Roman Urdu has also been proposed.}
\label{table:datasets}
\end{table*}

Most ground breaking research nowadays focuses on Multilingual language modelling \cite{conneau2019cross,conneau2019unsupervised,pires-etal-2019-multilingual,khawaja2018domain,
beg2008critical, beg2001memory, farooq2019melta}. This is also desirable from an industrial perspective where most app makers would like to easily scale to a global audiences. However a lot of research is also being done on pretraining large monolingual language models for each language 

\cite{virtanen2019multilingual,de2019bertje,baly2020arabert,yu2019adaptation,polignano2019alberto,canete2020spanish,martin2019camembert,le2020flaubert,rani2015case, beg2009flecs, beg2007flecs, koleilat2006watagent}. This can be desirable for achieving better and faster performance as compared to cross and multilingual language modeling. It has also been shown by recent works that monolingual models tend to outperform their multilingual counterparts of the same size in similar settings \cite{de2019bertje,martin2019camembert,virtanen2019multilingual,pyysalo2020wikibert,farooq2019bigdata,zafar2019constructive, zafar2018deceptive,thaver2016pulmonary}. Multilingual is also not always desirable, like when interested only in improving performance for a specific language. This is because Multilingual models suffer from the capacity dilution problem \cite{arivazhagan2019massively} also termed as the \textit{curse of multilinguality} \cite{conneau2019unsupervised}. This is can be handled by increasing model capacity but at the cost of memory and inference time. As the number of languages increase this leaves lesser space for each language.

Roman Urdu is a code-switched language \cite{beg2006maxsm} formed by a mixture of Urdu and English. It shares the vocabulary and uses the same Latin script as English and other words are transliterated from Urdu \cite{alvi2017ensights,zafar2020search}. It is widely used on many social media platforms and chat apps mainly in South Asian countries like India and Pakistan.  However it is a resource starved language as there is not a single corpus, tools or techniques to create Large Pretrained Language models and enable out-of-the-box Natural Language Processing (NLP) tasks \cite{beg2001memory}. In this research we take a bilingual language modeling approach instead of a multilingual one for Roman Urdu and show that its code-switching property with English can be intelligently used to obtain a better performance as compared to Mono and Multi lingual models. The main contributions of this work are as follows:
\begin{enumerate}
    \item We show how performance of a specific low-resource language can be improved by taking advantage of its code-switching property and a typo-logically similar high-resource language.
    \item We introduce a new method for producing Bilingual models from Monolingual models using additional pretraining.
    \item We apply the proposed Bilingual Language modeling techniques to Roman Urdu and show that significant performance gains can be achieved as compared to Monolingual modeling. 
    \item We also show how languages can be added to the linguistic space of Mono and Multilingual models by using vocabulary augmentation and additional pretraining.
    \item We apply the proposed Bilingual Language modeling techniques to Roman Urdu and show that significant performance gains can be achieved as compared to Monolingual modeling
    \item We propose a novel collection of Roman Urdu corpora gathered from various sources and also transliterated from Urdu.
    \item We make the corpora and pretrained Mono, Bi and Multi lingual BERT and RoBERTa models publicly available.
\end{enumerate}

The paper is organized as follows. In section \ref{s:2} we explain in detail the pretraining methodology and dataset used. Section \ref{s:3} discusses the evaluation methodology. Finally the Related work is discussed in section \ref{s:4}.

\begin{figure*}[htbp]
\centerline{\includegraphics[scale=.6]{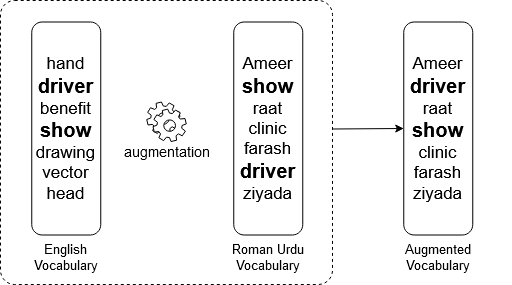}}
\caption{An example of vocabulary augmentation. Notice how the words \textit{driver} and \textit{show} retain positions in augmented vocabulary from English.}
\label{fig:vocabulary}
\end{figure*}

\section{Bilingual Language Modeling}
\label{s:2}

\subsection{Vocabulary Augmentation}
\label{technique}

The main contribution of this paper which enables the cross-lingual transfer learning \cite{baigahmed} involves vocabulary augmentation. This involves changing the existing vocabulary of a pretrained model using a specific criteria. After the vocabulary augmentation additional cycles of pretraining are carried out to produce the final model.

\begin{figure*}[htbp]
\centerline{\includegraphics[scale=.6]{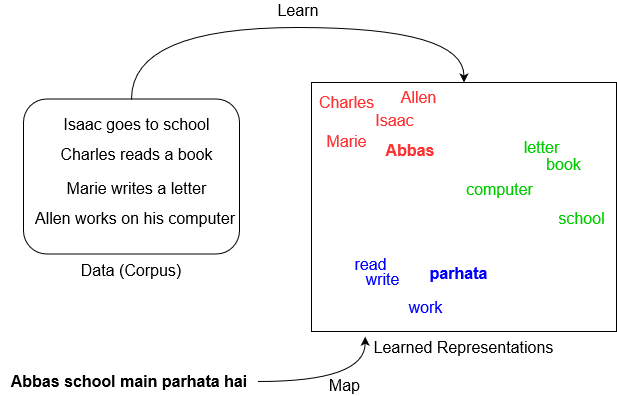}}
\caption{An abstract representation of mapping new Roman Urdu sentences to an existing English representation space.}
\label{fig:methodology}
\end{figure*}

The vocabulary augmentation technique \cite{seth2006achieving} can in theory be applied to any low resource language which shares vocabulary with a high resource language \cite{awan2021top}. In this research we apply this technique to Roman Urdu, a language which shares a significant amount of vocabulary \cite{javed2019fairness} with English and show that we achieve significantly improved performance. Roman Urdu uses the same Latin script as English and many other Languages in addition to this Roman Urdu also consists of many English words such that these are often used interchangeably \cite{beg2013constraint} in a code switched mode \cite{bangash2017methodology}. This opens up exciting opportunities when building a Roman Urdu Language model. An abstract representation of this idea is also shown in Figure \ref{fig:methodology}. Consider the sentence \mbox{\textbf{Abbas school main parhata hai}} (\textit{Abbas teaches in a school}). In this sentence the English model knows the context of \textit{school}. Thus is able to learn better word representations for the surrounding words as compared to training the model from scratch \cite{naeem2020deep}.

The advantage of this cross-lingual transfer \cite{sahar2019towards} is that it will enable a significantly improved performance for Roman Urdu without the need of extensive data and training cycles. An example of augmenting a small set of Roman Urdu vocabulary with English \cite{qamarrelationship} is shown in Fig. \ref{fig:vocabulary}. The words \textit{driver} and \textit{show} are common in both languages thus their positions from the original language (English) are retained.

The vocabulary augmentation method \cite{uzair2019weec} can be extended to any language which satisfy certain conditions. The augmentation method and its are described in detail as follows. Consider two languages $x$ and $y$ where $x$ is a low resource language and $y$ is high resource. Vocabulary augmentation can be applied for $x$ if equations \ref{eqn1} and \ref{eqn2} hold true. $L(x)$ is the super set of all the words of language $x$ and produces the vocabulary vector $V(x) = [x_1,x_2,x_3,...,x_n]$. Consider the vocabulary vector $V(y) = [y_1,y_2,y_3,...,y_n]$ given $L(x) \cap L(y) = \{y_1,y_3\}$, the augmented vocabulary vector $V(z)$ will be produced as follows $V(z) = [y_1,x_2,y_3,x_4,...,x_n]$. The positions of the common elements between $V(x)$ and $V(y)$ will not be changed while the other elements from $V(x)$ will be added without changing their position to $V(z)$.

\begin{equation}
\label{eqn1}
L(x)  \cap  L(y)  \neq  \emptyset
\end{equation}

\begin{equation}
\label{eqn2}
L(y) \subset L(x)
\end{equation}

\subsection{Training and Architecture}
This section explains the steps to transform data \cite{zafar2019using} for pretraining and the changes made to the pretraining cycles to enable cross-lingual transfer from English to Urdu. We also discuss the architectures used to perform pretraining for Mono, Bi and Multilingual models.

\textbf{Architecture.}
The architectures used in this research are mainly based on BERT and RoBERTa. All types of pretraining cycles use the BASE architecture \cite{beg2010graph} which consists of 12 layers, 768 hidden nodes, 12 attention heads and 110M parameters. The pretraining tasks of MLM (Masked Language Modeling) \cite{dilawar2018understanding}, NSP (Next Sentence Prediction) \cite{javed2020collaborative} and SOP (Sentence Order Prediction) use uncased vocabulary. These Uncased models typically have better performance overall. However cased versions are useful \cite{asad2020deepdetect} for tasks such as Part-of-Speech tagging or named entity recognition where the case of a letter encodes useful information.

\textbf{Pretraining.} Multilingual models \cite{karsten2007axiomatic} such as XLM and XLM-R have been trained on hundreds of languages of which Roman Urdu is not a part of. One way to solve this is using cross-lingual transfer which involves fine-tuning a Multilingual model on a specific Roman Urdu task or zero-shot cross-lingual transfer \cite{beg2019algorithmic} where model is fine-tuned on an English task and tested on Roman Urdu tasks. However this type of cross-lingual transfer is bound to yield a higher perplexity as the model has seen very less examples of Roman Urdu and the representation space is shared with many other languages. 

To overcome both these issues we perform cross-lingual transfer \cite{javed2020alphalogger} during pretraining phase using vocabulary augmentation and cut down the languages to a minimal. So the model's representation space only contains the high resource language the transfer is being made from and the low resource language to which the transfer is being made. To enable the cross-lingual transfer during training phase we run  additional pretraining cycles. This involves running some steps of the pretraining phase for the new language after augmenting Roman Urdu vocabulary with English. 

This notion of cross-lingual transfer \cite{beg2006performance} can also be applied to Multilingual models however the learning space is shared with many other languages and \textit{the curse of multilinguality} comes into play. These hypothesis are confirmed by the three types of pretraining Experiments we perform. The first type involves training from scratch. The second type involves additional pretraining of a Monolingual English model and the third type involves additional pretraining of a multilingual model. The results of these experiments are discussed in detail in Section \ref{s:3}.

\subsection{Pretraining Dataset}
\label{dataset}
This section explains the data gathering process and the preprocessing steps applied to transform the data in a suitable form as expected by the Byte Pair Encoder (BPE).

\textbf{Collection.} Roman Urdu as mentioned earlier is a resource starved language, little \cite{zahid2020roman,majeed2020emotion,khawaja2018domain} or no work has been done for Roman Urdu and almost no large publicly available dataset exists. Therefore to enable pretraining of Large Transformer based language models we propose a novel collection of corpora cleaned and transliterated to Roman Urdu \cite{tariq2019accurate}. In addition to this we also propose a novel Roman Urdu dataset consisting of 3 million Roman Urdu sentences and 54 million tokens. This dataset has been scraped from twitter where tweets contain at-least a small amount of Roman Urdu. The statistics of the Twitter scraped dataset and all other datasets is shown in Table \ref{table:datasets}. All the above mentioned datasets have also been made publicly available.

\textbf{Transliteration.} Most of the collected datasets are taken from prominent research works on Urdu while a few were taken from research works on Roman Urdu \cite{sharf2018performing,arshad2019corpus}. The collected Urdu datasets \newcite{khananamed,sharjeel2017counter,amjad2020bend,azam2020sentiment,nacem2020subspace} have then been transliterated to Roman Urdu. The transliteration process involves changing the Urdu Devanagri script to the Latin script used by Roman Urdu. All of the transliteration is performed using ijunoon's Urdu to Roman Urdu transliteration API \footnote{\url{https://www.ijunoon.com/transliteration/urdu-to-roman/}}.

\section{Experiments}
\label{s:3}

In this section we describe the methods used for evaluation and discuss the results obtained for the three different types of pretraining.

\begin{table*}
\centering
\begin{tabular}{lll}
\hline
\textbf{Model} & \textbf{MLM} & \textbf{NSP}\\
\hline
BERT\textsubscript{Monolingual} & 0.02 & 0.53 \\
BERT\textsubscript{Multilingual} & 0.11 & \textbf{0.91} \\
BERT\textsubscript{Bilingual} & \textbf{0.23} & \textbf{0.95} \\
\hline
\end{tabular}
\caption{ A comparison of MLM and NSP accuracies for three types of BERT models. As proposed the Bilingual model achieves the highest performance. }
\label{table:bert}
\end{table*}

\subsection{Evaluation} 

Large Pretrained language models are usually evaluated on a range of NLP tasks by fine tuning the model on each of the specific task. Previously BERT and BERT like models have been evaluated on the General Language Understanding Evaluation (GLUE) benchmark \cite{wang2018glue} which consists of a range of Natural Language Understanding (NLU) tasks. The tasks range from simple semantic analysis, textual entailment to complex reading comprehension tasks such as in the SQuAD \cite{rajpurkar2016squad,rajpurkar2018know} and QNLI \cite{wang2018glue} tasks. Other stricter evaluation tasks have also been proposed over time which consist of the SQuAD v2.0 \cite{rajpurkar2018know} and the SWAG \cite{zellers2018swag} datasets.

As research progressed, larger and better models like GPT \cite{radford2018improving,radford2019language}, XLNET \cite{yang2019xlnet} achieved close to human level performance on the GLUE benchmark. This lead to the introduction of SuperGLUE \cite{wang2019superglue} a set of more challenging tasks for producing better language models. However these evaluation tasks were only for English models and Multilingual or Monolingual models in other languages could not be evaluated. As the trend towards multilingual and universal language modeling increased this lead to the creation of Multi-task and Multi-lingual benchmarks like Xtreme \cite{hu2020xtreme} and XGLUE \cite{liang2020xglue}. Despite of the abundance of Multilingual and Multitask evaluation sets covering hundreds of languages no dataset provides support for Roman Urdu. Thus for evaluation of the performance of our pretrained models we use the validation metrics calculated during the pretraining phase of each model. The validation dataset is based on data that is put aside during the creation of pretraining data.

The Evaluation metrics include two tasks for BERT based models namely Masked Language Modeling (MLM) and Next Sentence Prediction (NSP) while the RoBERTa models are evaluated in terms of MLM loss and MLM perplexity. The Masked Language modeling task involves predicting a randomly masked word. A word is masked randomly about 15\% of the time for each given sentence. The Next Sentence prediction task involves predicting whether two given sentences are consecutive in the correct order i.e. sentence 2 comes after sentence 1. The MLM perplexity is calculated using eqn. \ref{eqn3}. This represents how sure the model is when predicting the masked word.

\begin{equation}
\label{eqn3}
PP(W) = \frac{1}{P(w_1,w_2,...,w_N)^{\frac{1}{n}}}
\end{equation}

\begin{table*}
\centering
\begin{tabular}{lllll}
\hline
\textbf{Model} & \textbf{Train loss} & \textbf{Valid loss} & \textbf{Train perplexity} & \textbf{Valid perplexity}\\
\hline
RoBERTa\textsubscript{Monolingual} & 2.01 & 2.46 & 4.08 & 5.49 \\
RoBERTa\textsubscript{Bilingual} & \textbf{1.72} & \textbf{2.19} & \textbf{3.24} & \textbf{4.56} \\
\hline
\end{tabular}
\caption{ Loss and perplexity for the MLM pretraining task. The proposed Bilingual modeling technique is able to outperform in both training and validation. }
\label{table:roberta}
\end{table*}

\subsection{Results} 

The evaluation of all BERT based models as mentioned before is performed using Masked Language Modeling (MLM) and Next Sentence Prediction (NSP) while the evaluation of RoBERTa models is done using Loss and Perplexity for the MLM task. \mbox{Table \ref{table:bert}} shows the accuracies of MLM and NSP tasks for the Monolingual, Bilingual and Multilingual BERT based models that we have trained.

The monolingual BERT is pretrained from scratch purely on Roman Urdu while the Multilingual BERT is trained by performing additional pretraining steps on a pretrained Multilingual BERT. The Bilingual BERT which achieves the best performance is trained by performing the same additional pretraining steps as before but on a monolingual English BERT Base model by following the Vocabulary Augmentation technique as described in section \ref{technique}. All pretraining cycles use the same novel corpora proposed in Section \ref{dataset}. Each of the pretraining and additional pretraining cycles are performed for 10K steps with a learning rate of $2e-5$ except for Monolingual training which was performed with a learning rate of $1e-4$, the higher learning rate was necessary when training from scratch. 

The Monolingual RoBERTa is trained from scratch purely on the proposed Roman Urdu datasets. The Bilingual RoBERTa is trained by starting from a pretrained English model and vocabulary augmentation is again used to fuse the new language to create a bilingual model. The evaluation results of RoBERTa based models are shown in Table \ref{table:roberta}. It can be seen that the Bilingual modeling approach is also able to outperform here. 

From the above experiments we see that Bilingual models of Roman Urdu significantly outperform their Monolingual and Multilingual counter parts in similar settings. This performance is significant in the MLM task as compared to the NSP task which also reiterates the idea in \newcite{liu2019roberta} that NSP is an easier task compared to MLM. The experiments conclude that the proposed vocabulary augmentation technique can be effectively used to significantly improve performance for Low Resource languages given that they share vocabulary with a high resource language.

\section{Related Work}
\label{s:4}

Neural Language modeling was first proposed by \cite{bengio2003neural,collobert2008unified} who showed that the model implicitly learnt useful representations. The technique popularly came to be known as word embeddings. These embeddings were a leap forward in the field of NLP notably after the introduction of techniques like word2vec \cite{mikolov2013distributed}, GloVe \cite{pennington2014glove}, fastText \cite{joulin2017bag}. These early techniques were mostly context-free and a major shortcoming was that they couldn’t handle Polysemy. This started the search for contextual embeddings. ELMo \cite{peters2018deep} and ULMFiT \cite{howard2018universal} were the first to achieve substantial  improvements using  LSTM  based  language models. Following the line of work of contextual models, a different approach, GPT \cite{radford2018improving} was proposed to tackle tasks in the GLUE benchmark \cite{wang2018glue}. The new approach to contextual language modelling was to replace LSTMs entirely with then recently released Transformers \cite{vaswani2017attention}. GPT used a 12-layer decoder-only transformer which was trained for 100 epochs on the BookCorpus \cite{zhu2015aligning}  and  achieved  considerable improvements as compared to ELMo on the same NLP  tasks. The  hugely  popular  BERT was based on a similar strategy as GPT containing  12  levels  but  only  used  the  encoder part of the Transformer and looked at sentences bidirectionally.

Many studies have been performed in the area of bilingual and multilingual language modeling. Several works focus on mapping word representations in multiple languages to a unified embedding space so words in different languages can be compared. One line of work focuses on using bilingually aligned corpora at sentence level \cite{zou2013bilingual,hermann2014multilingual,luong2015bilingual} and another tries to achieves the same on document level \cite{vulic2016bilingual,levy2016strong}. Another line of work focuses on utilising the isomorphic structure of languages, dictionary mappings and shared vocabulary for ad-hoc mappings between languages \cite{artetxe2018robust,faruqui2014improving}.

\section{Conclusion} 
In this research we propose a large Roman Urdu corpus with 3M sentences in addition to a novel collection of corpora cleaned and transliterated from Urdu. The corpora cover a variety of domains from news, IMDB reviews to tweets. In addition to this we introduce a novel vocabulary augmentation method that enables adding new languages to pretrained models. We also show how this new method can be used for cross-lingual transfer and improving performance for Low Resource languages. We have also shown that the proposed Bilingual language modeling is able to outperform multilingual modeling given a code-switched language like Roman Urdu. In the future we would like to better evaluate these Roman Urdu Language models on downstream NLP tasks which can be done by producing a GLUE like benchmark for Roman Urdu. We hope that the corpora and methods proposed in this paper will enable the NLP community to produce better language models especially for Low resource and resource starved Languages like Roman Urdu.

\bibliographystyle{acl_natbib}
\bibliography{anthology,acl2021}

\begin{thebibliography}{84}
\expandafter\ifx\csname natexlab\endcsname\relax\def\natexlab#1{#1}\fi

\bibitem[{Alvi et~al.(2017)Alvi, Sahar, Bangash, and Beg}]{alvi2017ensights}
Hamza~M Alvi, Hareem Sahar, Abdul~A Bangash, and Mirza~O Beg. 2017.
\newblock Ensights: A tool for energy aware software development.
\newblock In \emph{2017 13th International Conference on Emerging Technologies
  (ICET)}, pages 1--6. IEEE.

\bibitem[{Amjad et~al.(2020)Amjad, Sidorov, Zhila, G{\'o}mez-Adorno, Voronkov,
  and Gelbukh}]{amjad2020bend}
Maaz Amjad, Grigori Sidorov, Alisa Zhila, Helena G{\'o}mez-Adorno, Ilia
  Voronkov, and Alexander Gelbukh. 2020.
\newblock “bend the truth”: Benchmark dataset for fake news detection in
  urdu language and its evaluation.
\newblock \emph{Journal of Intelligent \& Fuzzy Systems}, 51(Preprint):1--13.

\bibitem[{Arivazhagan et~al.(2019)Arivazhagan, Bapna, Firat, Lepikhin, Johnson,
  Krikun, Chen, Cao, Foster, Cherry et~al.}]{arivazhagan2019massively}
Naveen Arivazhagan, Ankur Bapna, Orhan Firat, Dmitry Lepikhin, Melvin Johnson,
  Maxim Krikun, Mia~Xu Chen, Yuan Cao, George Foster, Colin Cherry, et~al.
  2019.
\newblock Massively multilingual neural machine translation in the wild:
  Findings and challenges.
\newblock \emph{arXiv preprint arXiv:1907.05019}.

\bibitem[{Arshad et~al.(2019)Arshad, Bashir, Majeed, Shahzad, and
  Beg}]{arshad2019corpus}
Muhammad~Umair Arshad, Muhammad~Farrukh Bashir, Adil Majeed, Waseem Shahzad,
  and Mirza~Omer Beg. 2019.
\newblock Corpus for emotion detection on roman urdu.
\newblock In \emph{2019 22nd International Multitopic Conference (INMIC)},
  pages 1--6. IEEE.

\bibitem[{Artetxe et~al.(2018)Artetxe, Labaka, and Agirre}]{artetxe2018robust}
Mikel Artetxe, Gorka Labaka, and Eneko Agirre. 2018.
\newblock A robust self-learning method for fully unsupervised cross-lingual
  mappings of word embeddings.
\newblock \emph{arXiv preprint arXiv:1805.06297}.

\bibitem[{Asad et~al.(2020)Asad, Asim, Javed, Beg, Mujtaba, and
  Abbas}]{asad2020deepdetect}
Muhammad Asad, Muhammad Asim, Talha Javed, Mirza~O Beg, Hasan Mujtaba, and
  Sohail Abbas. 2020.
\newblock Deepdetect: detection of distributed denial of service attacks using
  deep learning.
\newblock \emph{The Computer Journal}, 63(7):983--994.

\bibitem[{Awan and Beg(2021)}]{awan2021top}
Mubashar~Nazar Awan and Mirza~Omer Beg. 2021.
\newblock Top-rank: a topicalpostionrank for extraction and classification of
  keyphrases in text.
\newblock \emph{Computer Speech \& Language}, 65:101116.

\bibitem[{Azam et~al.(2020)Azam, Tahir, and Mehmood}]{azam2020sentiment}
Nazish Azam, Bilal Tahir, and Muhammad~Amir Mehmood. 2020.
\newblock Sentiment and emotion analysis of text: A survey on approaches and
  resources.
\newblock \emph{LANGUAGE \& TECHNOLOGY}, page~87.

\bibitem[{Baig et~al.()Baig, Beg, Bhatti, Bhuiyan, Bissyand{\'e}, Chen,
  Chhetri, Couto, de~Macedo, de~Vries et~al.}]{baigahmed}
Zubair Baig, Mirza~Omer Beg, Baber~Majid Bhatti, Farzana~Ahamed Bhuiyan,
  Tegawend{\'e}~F Bissyand{\'e}, Shizhan Chen, Mohan~Baruwal Chhetri, Marco
  Couto, Jo{\~a}o de~Macedo, Randy de~Vries, et~al.
\newblock Ahmed, sanam 124 aleti, aldeida 105 alo{\'\i}sio, jo{\~a}o 151
  arachchilage, nalin asanka gamagedara 7.

\bibitem[{Baly et~al.(2020)Baly, Hajj et~al.}]{baly2020arabert}
Fady Baly, Hazem Hajj, et~al. 2020.
\newblock Arabert: Transformer-based model for arabic language understanding.
\newblock In \emph{Proceedings of the 4th Workshop on Open-Source Arabic
  Corpora and Processing Tools, with a Shared Task on Offensive Language
  Detection}, pages 9--15.

\bibitem[{Bangash et~al.(2017)Bangash, Sahar, and Beg}]{bangash2017methodology}
Abdul~Ali Bangash, Hareem Sahar, and Mirza~Omer Beg. 2017.
\newblock A methodology for relating software structure with energy
  consumption.
\newblock In \emph{2017 IEEE 17th International Working Conference on Source
  Code Analysis and Manipulation (SCAM)}, pages 111--120. IEEE.

\bibitem[{Beg(2008)}]{beg2008critical}
M~Beg. 2008.
\newblock Critical path heuristic for automatic parallelization.

\bibitem[{Beg(2009)}]{beg2009flecs}
Mirza Beg. 2009.
\newblock Flecs: A framework for rapidly implementing forwarding protocols.
\newblock In \emph{International Conference on Complex Sciences}, pages
  1761--1773. Springer.

\bibitem[{Beg and Beek(2013)}]{beg2013constraint}
Mirza Beg and Peter~van Beek. 2013.
\newblock A constraint programming approach for integrated spatial and temporal
  scheduling for clustered architectures.
\newblock \emph{ACM Transactions on Embedded Computing Systems (TECS)},
  13(1):1--23.

\bibitem[{Beg et~al.(2006)Beg, Charlin, and So}]{beg2006maxsm}
Mirza Beg, Laurent Charlin, and Joel So. 2006.
\newblock Maxsm: A multi-heuristic approach to xml schema matching.

\bibitem[{Beg and Dahlin()}]{beg2001memory}
Mirza Beg and Mike Dahlin.
\newblock A memory accounting interface for the java programming language.

\bibitem[{Beg and Van~Beek(2010)}]{beg2010graph}
Mirza Beg and Peter Van~Beek. 2010.
\newblock A graph theoretic approach to cache-conscious placement of data for
  direct mapped caches.
\newblock In \emph{Proceedings of the 2010 international symposium on Memory
  management}, pages 113--120.

\bibitem[{Beg et~al.(2019)Beg, Awan, and Ali}]{beg2019algorithmic}
Mirza~O Beg, Mubashar~Nazar Awan, and Syed~Shahzaib Ali. 2019.
\newblock Algorithmic machine learning for prediction of stock prices.
\newblock In \emph{FinTech as a Disruptive Technology for Financial
  Institutions}, pages 142--169. IGI Global.

\bibitem[{Beg(2006)}]{beg2006performance}
Mirza~Omer Beg. 2006.
\newblock Performance analysis of packet forwarding on ixp2400 network
  processor.

\bibitem[{Beg(2007)}]{beg2007flecs}
Mirza~Omer Beg. 2007.
\newblock Flecs: A data-driven framework for rapid protocol prototyping.
\newblock Master's thesis, University of Waterloo.

\bibitem[{Bengio et~al.(2003)Bengio, Ducharme, Vincent, and
  Jauvin}]{bengio2003neural}
Yoshua Bengio, R{\'e}jean Ducharme, Pascal Vincent, and Christian Jauvin. 2003.
\newblock A neural probabilistic language model.
\newblock \emph{Journal of machine learning research}, 3(Feb):1137--1155.

\bibitem[{Canete et~al.(2020)Canete, Chaperon, Fuentes, and
  P{\'e}rez}]{canete2020spanish}
Jos{\'e} Canete, Gabriel Chaperon, Rodrigo Fuentes, and Jorge P{\'e}rez. 2020.
\newblock Spanish pre-trained bert model and evaluation data.
\newblock \emph{PML4DC at ICLR}, 2020.

\bibitem[{Collobert and Weston(2008)}]{collobert2008unified}
Ronan Collobert and Jason Weston. 2008.
\newblock A unified architecture for natural language processing: Deep neural
  networks with multitask learning.
\newblock In \emph{Proceedings of the 25th international conference on Machine
  learning}, pages 160--167.

\bibitem[{Conneau et~al.(2020)Conneau, Khandelwal, Goyal, Chaudhary, Wenzek,
  Guzm{\'{a}}n, Grave, Ott, Zettlemoyer, and
  Stoyanov}]{conneau2019unsupervised}
Alexis Conneau, Kartikay Khandelwal, Naman Goyal, Vishrav Chaudhary, Guillaume
  Wenzek, Francisco Guzm{\'{a}}n, Edouard Grave, Myle Ott, Luke Zettlemoyer,
  and Veselin Stoyanov. 2020.
\newblock \href {https://www.aclweb.org/anthology/2020.acl-main.747/}
  {Unsupervised cross-lingual representation learning at scale}.
\newblock In \emph{Proceedings of the 58th Annual Meeting of the Association
  for Computational Linguistics, {ACL} 2020, Online, July 5-10, 2020}, pages
  8440--8451. Association for Computational Linguistics.

\bibitem[{Conneau and Lample(2019)}]{conneau2019cross}
Alexis Conneau and Guillaume Lample. 2019.
\newblock Cross-lingual language model pretraining.
\newblock In \emph{Advances in Neural Information Processing Systems}, pages
  7059--7069.

\bibitem[{Dilawar et~al.(2018)Dilawar, Majeed, Beg, Ejaz, Muhammad, Mehmood,
  and Nam}]{dilawar2018understanding}
Noman Dilawar, Hammad Majeed, Mirza~Omer Beg, Naveed Ejaz, Khan Muhammad, Irfan
  Mehmood, and Yunyoung Nam. 2018.
\newblock Understanding citizen issues through reviews: A step towards data
  informed planning in smart cities.
\newblock \emph{Applied Sciences}, 8(9):1589.

\bibitem[{Farooq et~al.(2019{\natexlab{a}})Farooq, Beg
  et~al.}]{farooq2019bigdata}
Muhammad~Umer Farooq, Mirza~Omer Beg, et~al. 2019{\natexlab{a}}.
\newblock Bigdata analysis of stack overflow for energy consumption of android
  framework.
\newblock In \emph{2019 International Conference on Innovative Computing
  (ICIC)}, pages 1--9. IEEE.

\bibitem[{Farooq et~al.(2019{\natexlab{b}})Farooq, Khan, and
  Beg}]{farooq2019melta}
Muhammad~Umer Farooq, Saif Ur~Rehman Khan, and Mirza~Omer Beg.
  2019{\natexlab{b}}.
\newblock Melta: A method level energy estimation technique for android
  development.
\newblock In \emph{2019 International Conference on Innovative Computing
  (ICIC)}, pages 1--10. IEEE.

\bibitem[{Faruqui and Dyer(2014)}]{faruqui2014improving}
Manaal Faruqui and Chris Dyer. 2014.
\newblock Improving vector space word representations using multilingual
  correlation.
\newblock In \emph{Proceedings of the 14th Conference of the European Chapter
  of the Association for Computational Linguistics}, pages 462--471.

\bibitem[{Hermann and Blunsom(2014)}]{hermann2014multilingual}
Karl~Moritz Hermann and Phil Blunsom. 2014.
\newblock Multilingual models for compositional distributed semantics.
\newblock \emph{arXiv preprint arXiv:1404.4641}.

\bibitem[{Howard and Ruder(2018)}]{howard2018universal}
Jeremy Howard and Sebastian Ruder. 2018.
\newblock Universal language model fine-tuning for text classification.
\newblock In \emph{Proceedings of the 56th Annual Meeting of the Association
  for Computational Linguistics (Volume 1: Long Papers)}, pages 328--339.

\bibitem[{Hu et~al.(2020)Hu, Ruder, Siddhant, Neubig, Firat, and
  Johnson}]{hu2020xtreme}
Junjie Hu, Sebastian Ruder, Aditya Siddhant, Graham Neubig, Orhan Firat, and
  Melvin Johnson. 2020.
\newblock Xtreme: A massively multilingual multi-task benchmark for evaluating
  cross-lingual generalization.
\newblock \emph{arXiv preprint arXiv:2003.11080}.

\bibitem[{Javed et~al.(2020{\natexlab{a}})Javed, Beg, Asim, Baker, and
  Al-Bayatti}]{javed2020alphalogger}
Abdul~Rehman Javed, Mirza~Omer Beg, Muhammad Asim, Thar Baker, and Ali~Hilal
  Al-Bayatti. 2020{\natexlab{a}}.
\newblock Alphalogger: Detecting motion-based side-channel attack using
  smartphone keystrokes.
\newblock \emph{Journal of Ambient Intelligence and Humanized Computing}, pages
  1--14.

\bibitem[{Javed et~al.(2020{\natexlab{b}})Javed, Sarwar, Beg, Asim, Baker, and
  Tawfik}]{javed2020collaborative}
Abdul~Rehman Javed, Muhammad~Usman Sarwar, Mirza~Omer Beg, Muhammad Asim, Thar
  Baker, and Hissam Tawfik. 2020{\natexlab{b}}.
\newblock A collaborative healthcare framework for shared healthcare plan with
  ambient intelligence.
\newblock \emph{Human-centric Computing and Information Sciences}, 10(1):1--21.

\bibitem[{Javed et~al.(2019)Javed, Beg, Mujtaba, Majeed, and
  Asim}]{javed2019fairness}
Hafiz~Tayyeb Javed, Mirza~Omer Beg, Hasan Mujtaba, Hammad Majeed, and Muhammad
  Asim. 2019.
\newblock Fairness in real-time energy pricing for smart grid using
  unsupervised learning.
\newblock \emph{The Computer Journal}, 62(3):414--429.

\bibitem[{Joulin et~al.(2017)Joulin, Grave, Bojanowski, and
  Mikolov}]{joulin2017bag}
Armand Joulin, {\'E}douard Grave, Piotr Bojanowski, and Tom{\'a}{\v{s}}
  Mikolov. 2017.
\newblock Bag of tricks for efficient text classification.
\newblock In \emph{Proceedings of the 15th Conference of the European Chapter
  of the Association for Computational Linguistics: Volume 2, Short Papers},
  pages 427--431.

\bibitem[{Karsten et~al.(2007)Karsten, Keshav, Prasad, and
  Beg}]{karsten2007axiomatic}
Martin Karsten, Srinivasan Keshav, Sanjiva Prasad, and Mirza Beg. 2007.
\newblock An axiomatic basis for communication.
\newblock \emph{ACM SIGCOMM Computer Communication Review}, 37(4):217--228.

\bibitem[{Khana et~al.(2016)Khana, Daudb, Nasira, and Amjada}]{khananamed}
Wahab Khana, Ali Daudb, Jamal~A Nasira, and Tehmina Amjada. 2016.
\newblock Named entity dataset for urdu named entity recognition task.
\newblock \emph{LANGUAGE \& TECHNOLOGY}, page~51.

\bibitem[{Khawaja et~al.(2018)Khawaja, Beg, and Qamar}]{khawaja2018domain}
Hussain~S Khawaja, Mirza~O Beg, and Saira Qamar. 2018.
\newblock Domain specific emotion lexicon expansion.
\newblock In \emph{2018 14th International Conference on Emerging Technologies
  (ICET)}, pages 1--5. IEEE.

\bibitem[{Koleilat et~al.(2006)Koleilat, So, and Beg}]{koleilat2006watagent}
Walid Koleilat, Joel So, and Mirza Beg. 2006.
\newblock Watagent: A fresh look at tac-scm agent design.

\bibitem[{Le et~al.(2020)Le, Vial, Frej, Segonne, Coavoux, Lecouteux, Allauzen,
  Crabb{\'e}, Besacier, and Schwab}]{le2020flaubert}
Hang Le, Lo{\"\i}c Vial, Jibril Frej, Vincent Segonne, Maximin Coavoux,
  Benjamin Lecouteux, Alexandre Allauzen, Benoit Crabb{\'e}, Laurent Besacier,
  and Didier Schwab. 2020.
\newblock Flaubert: Unsupervised language model pre-training for french.
\newblock In \emph{Proceedings of The 12th Language Resources and Evaluation
  Conference}, pages 2479--2490.

\bibitem[{Levy et~al.(2016)Levy, S{\o}gaard, and Goldberg}]{levy2016strong}
Omer Levy, Anders S{\o}gaard, and Yoav Goldberg. 2016.
\newblock A strong baseline for learning cross-lingual word embeddings from
  sentence alignments.
\newblock \emph{arXiv preprint arXiv:1608.05426}.

\bibitem[{Liang et~al.(2020)Liang, Duan, Gong, Wu, Guo, Qi, Gong, Shou, Jiang,
  Cao et~al.}]{liang2020xglue}
Yaobo Liang, Nan Duan, Yeyun Gong, Ning Wu, Fenfei Guo, Weizhen Qi, Ming Gong,
  Linjun Shou, Daxin Jiang, Guihong Cao, et~al. 2020.
\newblock Xglue: A new benchmark dataset for cross-lingual pre-training,
  understanding and generation.
\newblock \emph{arXiv preprint arXiv:2004.01401}.

\bibitem[{Liu et~al.(2019)Liu, Ott, Goyal, Du, Joshi, Chen, Levy, Lewis,
  Zettlemoyer, and Stoyanov}]{liu2019roberta}
Yinhan Liu, Myle Ott, Naman Goyal, Jingfei Du, Mandar Joshi, Danqi Chen, Omer
  Levy, Mike Lewis, Luke Zettlemoyer, and Veselin Stoyanov. 2019.
\newblock \href {http://arxiv.org/abs/1907.11692} {Roberta: {A} robustly
  optimized {BERT} pretraining approach}.
\newblock \emph{CoRR}, abs/1907.11692.

\bibitem[{Luong et~al.(2015)Luong, Pham, and Manning}]{luong2015bilingual}
Minh-Thang Luong, Hieu Pham, and Christopher~D Manning. 2015.
\newblock Bilingual word representations with monolingual quality in mind.
\newblock In \emph{Proceedings of the 1st Workshop on Vector Space Modeling for
  Natural Language Processing}, pages 151--159.

\bibitem[{Majeed et~al.(2020)Majeed, Mujtaba, and Beg}]{majeed2020emotion}
Adil Majeed, Hasan Mujtaba, and Mirza~Omer Beg. 2020.
\newblock Emotion detection in roman urdu text using machine learning.
\newblock In \emph{Proceedings of the 35th IEEE/ACM International Conference on
  Automated Software Engineering Workshops}, pages 125--130.

\bibitem[{Martin et~al.(2020)Martin, M{\"{u}}ller, Su{\'{a}}rez, Dupont,
  Romary, de~la Clergerie, Seddah, and Sagot}]{martin2019camembert}
Louis Martin, Benjamin M{\"{u}}ller, Pedro Javier~Ortiz Su{\'{a}}rez, Yoann
  Dupont, Laurent Romary, {\'{E}}ric de~la Clergerie, Djam{\'{e}} Seddah, and
  Beno{\^{\i}}t Sagot. 2020.
\newblock \href {https://www.aclweb.org/anthology/2020.acl-main.645/}
  {Camembert: a tasty french language model}.
\newblock In \emph{Proceedings of the 58th Annual Meeting of the Association
  for Computational Linguistics, {ACL} 2020, Online, July 5-10, 2020}, pages
  7203--7219. Association for Computational Linguistics.

\bibitem[{Mikolov et~al.(2013)Mikolov, Sutskever, Chen, Corrado, and
  Dean}]{mikolov2013distributed}
Tomas Mikolov, Ilya Sutskever, Kai Chen, Greg~S Corrado, and Jeff Dean. 2013.
\newblock Distributed representations of words and phrases and their
  compositionality.
\newblock In \emph{Advances in neural information processing systems (NIPS)},
  pages 3111--3119.

\bibitem[{Nacem et~al.(2020)Nacem, Iqbal, Saqib, Saad, Raza, Ali, Akhtar, Beg,
  Shahzad, and Arshad}]{nacem2020subspace}
Saad Nacem, Majid Iqbal, Muhammad Saqib, Muhammad Saad, Muhammad~Soban Raza,
  Zaid Ali, Naveed Akhtar, Mirza~Omer Beg, Waseem Shahzad, and Muhhamad~Umair
  Arshad. 2020.
\newblock Subspace gaussian mixture model for continuous urdu speech
  recognition using kaldi.
\newblock In \emph{2020 14th International Conference on Open Source Systems
  and Technologies (ICOSST)}, pages 1--7. IEEE.

\bibitem[{Naeem et~al.(2020)Naeem, Khan, Beg, and Mujtaba}]{naeem2020deep}
Bilal Naeem, Aymen Khan, Mirza~Omer Beg, and Hasan Mujtaba. 2020.
\newblock A deep learning framework for clickbait detection on social area
  network using natural language cues.
\newblock \emph{Journal of Computational Social Science}, pages 1--13.

\bibitem[{Pennington et~al.(2014)Pennington, Socher, and
  Manning}]{pennington2014glove}
Jeffrey Pennington, Richard Socher, and Christopher~D Manning. 2014.
\newblock Glove: Global vectors for word representation.
\newblock In \emph{Proceedings of the 2014 conference on empirical methods in
  natural language processing (EMNLP)}, pages 1532--1543.

\bibitem[{Peters et~al.(2018)Peters, Neumann, Iyyer, Gardner, Clark, Lee, and
  Zettlemoyer}]{peters2018deep}
Matthew~E Peters, Mark Neumann, Mohit Iyyer, Matt Gardner, Christopher Clark,
  Kenton Lee, and Luke Zettlemoyer. 2018.
\newblock Deep contextualized word representations.
\newblock In \emph{Proceedings of NAACL-HLT}, pages 2227--2237.

\bibitem[{Pires et~al.(2019)Pires, Schlinger, and
  Garrette}]{pires-etal-2019-multilingual}
Telmo Pires, Eva Schlinger, and Dan Garrette. 2019.
\newblock \href {https://doi.org/10.18653/v1/P19-1493} {How multilingual is
  multilingual {BERT}?}
\newblock In \emph{Proceedings of the 57th Annual Meeting of the Association
  for Computational Linguistics}, pages 4996--5001, Florence, Italy.
  Association for Computational Linguistics.

\bibitem[{Polignano et~al.(2019)Polignano, Basile, de~Gemmis, Semeraro, and
  Basile}]{polignano2019alberto}
Marco Polignano, Pierpaolo Basile, Marco de~Gemmis, Giovanni Semeraro, and
  Valerio Basile. 2019.
\newblock \href {http://ceur-ws.org/Vol-2481/paper57.pdf} {Alberto: Italian
  {BERT} language understanding model for {NLP} challenging tasks based on
  tweets}.
\newblock In \emph{Proceedings of the Sixth Italian Conference on Computational
  Linguistics, Bari, Italy, November 13-15, 2019}, volume 2481 of \emph{{CEUR}
  Workshop Proceedings}. CEUR-WS.org.

\bibitem[{Pyysalo et~al.(2020)Pyysalo, Kanerva, Virtanen, and
  Ginter}]{pyysalo2020wikibert}
Sampo Pyysalo, Jenna Kanerva, Antti Virtanen, and Filip Ginter. 2020.
\newblock \href {http://arxiv.org/abs/2006.01538} {Wikibert models: deep
  transfer learning for many languages}.
\newblock \emph{CoRR}, abs/2006.01538.

\bibitem[{Qamar et~al.()Qamar, Mujtaba, Majeed, and Beg}]{qamarrelationship}
Saira Qamar, Hasan Mujtaba, Hammad Majeed, and Mirza~Omer Beg.
\newblock Relationship identification between conversational agents using
  emotion analysis.
\newblock \emph{Cognitive Computation}, pages 1--15.

\bibitem[{Radford et~al.(2018)Radford, Narasimhan, Salimans, and
  Sutskever}]{radford2018improving}
Alec Radford, Karthik Narasimhan, Tim Salimans, and Ilya Sutskever. 2018.
\newblock Improving language understanding by generative pre-training.

\bibitem[{Radford et~al.(2019)Radford, Wu, Child, Luan, Amodei, and
  Sutskever}]{radford2019language}
Alec Radford, Jeffrey Wu, Rewon Child, David Luan, Dario Amodei, and Ilya
  Sutskever. 2019.
\newblock Language models are unsupervised multitask learners.
\newblock \emph{OpenAI blog}, 1(8):9.

\bibitem[{Rajpurkar et~al.(2018)Rajpurkar, Jia, and Liang}]{rajpurkar2018know}
Pranav Rajpurkar, Robin Jia, and Percy Liang. 2018.
\newblock Know what you don’t know: Unanswerable questions for squad.
\newblock In \emph{Proceedings of the 56th Annual Meeting of the Association
  for Computational Linguistics (Volume 2: Short Papers)}, pages 784--789.

\bibitem[{Rajpurkar et~al.(2016)Rajpurkar, Zhang, Lopyrev, and
  Liang}]{rajpurkar2016squad}
Pranav Rajpurkar, Jian Zhang, Konstantin Lopyrev, and Percy Liang. 2016.
\newblock Squad: 100,000+ questions for machine comprehension of text.
\newblock In \emph{Proceedings of the 2016 Conference on Empirical Methods in
  Natural Language Processing}, pages 2383--2392.

\bibitem[{Rani et~al.(2015)Rani, Imdad, and Beg}]{rani2015case}
Uzma Rani, Aamer Imdad, and Mirza Beg. 2015.
\newblock Case 2: Recurrent anemia in a 10-year-old girl.
\newblock \emph{Pediatrics in review}, 36(12):548--550.

\bibitem[{Sahar et~al.(2019)Sahar, Bangash, and Beg}]{sahar2019towards}
Hareem Sahar, Abdul~A Bangash, and Mirza~O Beg. 2019.
\newblock Towards energy aware object-oriented development of android
  applications.
\newblock \emph{Sustainable Computing: Informatics and Systems}, 21:28--46.

\bibitem[{Seth and Beg(2006)}]{seth2006achieving}
Aaditeshwar Seth and Mirza Beg. 2006.
\newblock Achieving privacy and security in radio frequency identification.
\newblock In \emph{Proceedings of the 2006 International Conference on Privacy,
  Security and Trust: Bridge the Gap Between PST Technologies and Business
  Services}, pages 1--1.

\bibitem[{Sharf and Rahman(2018)}]{sharf2018performing}
Zareen Sharf and Saif~Ur Rahman. 2018.
\newblock Performing natural language processing on roman urdu datasets.
\newblock \emph{International Journal Of Computer Science And Network
  Security}, 18(1):141--148.

\bibitem[{Sharjeel et~al.(2017)Sharjeel, Nawab, and
  Rayson}]{sharjeel2017counter}
Muhammad Sharjeel, Rao Muhammad~Adeel Nawab, and Paul Rayson. 2017.
\newblock Counter: corpus of urdu news text reuse.
\newblock \emph{Language resources and evaluation}, 51(3):777--803.

\bibitem[{Tariq et~al.(2019)Tariq, Majeed, Beg, Khan, and
  Derhab}]{tariq2019accurate}
Muhammad Tariq, Hammad Majeed, Mirza~Omer Beg, Farrukh~Aslam Khan, and
  Abdelouahid Derhab. 2019.
\newblock Accurate detection of sitting posture activities in a secure iot
  based assisted living environment.
\newblock \emph{Future Generation Computer Systems}, 92:745--757.

\bibitem[{Thaver and Beg(2016)}]{thaver2016pulmonary}
Danyal Thaver and Mirza Beg. 2016.
\newblock Pulmonary crohn's disease in down syndrome: A link or linkage
  problem.
\newblock \emph{Case reports in gastroenterology}, 10(2):206--211.

\bibitem[{Uzair et~al.(2019)Uzair, Beg, Mujtaba, and Majeed}]{uzair2019weec}
Ahmed Uzair, Mirza~O Beg, Hasan Mujtaba, and Hammad Majeed. 2019.
\newblock Weec: Web energy efficient computing: A machine learning approach.
\newblock \emph{Sustainable Computing: Informatics and Systems}, 22:230--243.

\bibitem[{Vaswani et~al.(2017)Vaswani, Shazeer, Parmar, Uszkoreit, Jones,
  Gomez, Kaiser, and Polosukhin}]{vaswani2017attention}
Ashish Vaswani, Noam Shazeer, Niki Parmar, Jakob Uszkoreit, Llion Jones,
  Aidan~N Gomez, {\L}ukasz Kaiser, and Illia Polosukhin. 2017.
\newblock Attention is all you need.
\newblock In \emph{Advances in neural information processing systems}, pages
  5998--6008.

\bibitem[{Virtanen et~al.(2019)Virtanen, Kanerva, Ilo, Luoma, Luotolahti,
  Salakoski, Ginter, and Pyysalo}]{virtanen2019multilingual}
Antti Virtanen, Jenna Kanerva, Rami Ilo, Jouni Luoma, Juhani Luotolahti, Tapio
  Salakoski, Filip Ginter, and Sampo Pyysalo. 2019.
\newblock Multilingual is not enough: Bert for finnish.
\newblock \emph{arXiv}, pages arXiv--1912.

\bibitem[{de~Vries et~al.(2019)de~Vries, van Cranenburgh, Bisazza, Caselli, van
  Noord, and Nissim}]{de2019bertje}
Wietse de~Vries, Andreas van Cranenburgh, Arianna Bisazza, Tommaso Caselli,
  Gertjan van Noord, and Malvina Nissim. 2019.
\newblock Bertje: A dutch bert model.
\newblock \emph{arXiv preprint arXiv:1912.09582}.

\bibitem[{Vuli{\'c} and Moens(2016)}]{vulic2016bilingual}
Ivan Vuli{\'c} and Marie-Francine Moens. 2016.
\newblock Bilingual distributed word representations from document-aligned
  comparable data.
\newblock \emph{Journal of Artificial Intelligence Research}, 55:953--994.

\bibitem[{Wang et~al.(2019)Wang, Pruksachatkun, Nangia, Singh, Michael, Hill,
  Levy, and Bowman}]{wang2019superglue}
Alex Wang, Yada Pruksachatkun, Nikita Nangia, Amanpreet Singh, Julian Michael,
  Felix Hill, Omer Levy, and Samuel Bowman. 2019.
\newblock Superglue: A stickier benchmark for general-purpose language
  understanding systems.
\newblock In \emph{Advances in neural information processing systems}, pages
  3266--3280.

\bibitem[{Wang et~al.(2018)Wang, Singh, Michael, Hill, Levy, and
  Bowman}]{wang2018glue}
Alex Wang, Amanpreet Singh, Julian Michael, Felix Hill, Omer Levy, and Samuel
  Bowman. 2018.
\newblock Glue: A multi-task benchmark and analysis platform for natural
  language understanding.
\newblock In \emph{Proceedings of the 2018 EMNLP Workshop BlackboxNLP:
  Analyzing and Interpreting Neural Networks for NLP}, pages 353--355.

\bibitem[{Yang et~al.(2019)Yang, Dai, Yang, Carbonell, Salakhutdinov, and
  Le}]{yang2019xlnet}
Zhilin Yang, Zihang Dai, Yiming Yang, Jaime Carbonell, Russ~R Salakhutdinov,
  and Quoc~V Le. 2019.
\newblock Xlnet: Generalized autoregressive pretraining for language
  understanding.
\newblock In \emph{Advances in neural information processing systems}, pages
  5753--5763.

\bibitem[{Yu and Arkhipov(2019)}]{yu2019adaptation}
Kuratov Yu and M~Arkhipov. 2019.
\newblock Adaptation of deep bidirectional multilingual transformers for
  russian language.
\newblock \emph{Computational Linguistics and Intellectual Technologies}, pages
  333--339.

\bibitem[{Zafar et~al.(2019{\natexlab{a}})Zafar, Mujtaba, Ashiq, and
  Beg}]{zafar2019constructive}
Adeel Zafar, Hasan Mujtaba, Sohrab Ashiq, and Mirza~Omer Beg.
  2019{\natexlab{a}}.
\newblock A constructive approach for general video game level generation.
\newblock In \emph{2019 11th Computer Science and Electronic Engineering
  (CEEC)}, pages 102--107. IEEE.

\bibitem[{Zafar et~al.(2019{\natexlab{b}})Zafar, Mujtaba, Baig, and
  Beg}]{zafar2019using}
Adeel Zafar, Hasan Mujtaba, Mirza~Tauseef Baig, and Mirza~Omer Beg.
  2019{\natexlab{b}}.
\newblock Using patterns as objectives for general video game level generation.
\newblock \emph{ICGA Journal}, 41(2):66--77.

\bibitem[{Zafar et~al.(2020)Zafar, Mujtaba, and Beg}]{zafar2020search}
Adeel Zafar, Hasan Mujtaba, and Mirza~Omer Beg. 2020.
\newblock Search-based procedural content generation for gvg-lg.
\newblock \emph{Applied Soft Computing}, 86:105909.

\bibitem[{Zafar et~al.(2018)Zafar, Mujtaba, Beg, and Ali}]{zafar2018deceptive}
Adeel Zafar, Hasan Mujtaba, Mirza~Omer Beg, and Sajid Ali. 2018.
\newblock Deceptive level generator.

\bibitem[{Zahid et~al.(2020)Zahid, Idrees, Mujtaba, and Beg}]{zahid2020roman}
Rabail Zahid, Muhammad~Owais Idrees, Hasan Mujtaba, and Mirza~Omer Beg. 2020.
\newblock Roman urdu reviews dataset for aspect based opinion mining.
\newblock In \emph{2020 35th IEEE/ACM International Conference on Automated
  Software Engineering Workshops (ASEW)}, pages 138--143. IEEE.

\bibitem[{Zellers et~al.(2018)Zellers, Bisk, Schwartz, and
  Choi}]{zellers2018swag}
Rowan Zellers, Yonatan Bisk, Roy Schwartz, and Yejin Choi. 2018.
\newblock Swag: A large-scale adversarial dataset for grounded commonsense
  inference.
\newblock In \emph{Proceedings of the 2018 Conference on Empirical Methods in
  Natural Language Processing}, pages 93--104.

\bibitem[{Zhu et~al.(2015)Zhu, Kiros, Zemel, Salakhutdinov, Urtasun, Torralba,
  and Fidler}]{zhu2015aligning}
Yukun Zhu, Ryan Kiros, Rich Zemel, Ruslan Salakhutdinov, Raquel Urtasun,
  Antonio Torralba, and Sanja Fidler. 2015.
\newblock Aligning books and movies: Towards story-like visual explanations by
  watching movies and reading books.
\newblock In \emph{Proceedings of the IEEE international conference on computer
  vision}, pages 19--27.

\bibitem[{Zou et~al.(2013)Zou, Socher, Cer, and Manning}]{zou2013bilingual}
Will~Y Zou, Richard Socher, Daniel Cer, and Christopher~D Manning. 2013.
\newblock Bilingual word embeddings for phrase-based machine translation.
\newblock In \emph{Proceedings of the 2013 Conference on Empirical Methods in
  Natural Language Processing}, pages 1393--1398.

\end{thebibliography}


\end{document}